\documentclass[letterpaper, 10 pt, conference]{ieeeconf}                         
\IEEEoverridecommandlockouts  
\usepackage{cite}
\usepackage[font=footnotesize]{subfig}
\usepackage{multirow}
\usepackage{comment}
\usepackage{ifpdf}
\usepackage[pdftex]{graphicx}
\usepackage{epstopdf}
\usepackage[cmex10]{amsmath}
\usepackage{algorithmic}
\usepackage{array}
\usepackage{mdwmath}
\usepackage{mdwtab}
\usepackage{eqparbox}
\usepackage{verbatim}
 \usepackage[ruled,vlined]{algorithm2e}
\usepackage[font=footnotesize]{subfig}
\usepackage{fixltx2e}
\usepackage{stfloats}
\usepackage{comment}
\usepackage{amsmath,bm}
\usepackage{amssymb}
\usepackage{color}
\usepackage{subfig}

\usepackage{multirow}
\usepackage{float}
\usepackage{caption}
\usepackage{comment}
\usepackage{hyperref}

\begin{document}
	
\title{\LARGE \bf
IoHRT: An Open-Source Unified Framework Towards the Internet of Humans and Robotic Things with Cloud Computing for Home-Care Applications}
\author{Dandan Zhang*, Jin Zheng*, Jialin Lin*
\thanks{All authors are with the Department of Engineering Mathematics, University of Bristol, Bristol, United Kingdom. Dandan Zhang and Jialin Lin are affiliated with Bristol Robotics Laboratory. }}

\maketitle

\begin{abstract}
The accelerating aging population has led to an increasing demand for domestic robotics to ease caregivers' burden. The integration of Internet of Things (IoT), robotics, and human-robot interaction (HRI) technologies is essential for home-care applications. Although the concept of the Internet of Robotic Things (IoRT) has been utilized in various fields, most existing IoRT frameworks lack ergonomic HRI interfaces and are limited to  specific tasks.

This paper presents an open-source unified Internet of Humans and Robotic Things (IoHRT) framework with cloud computing, which combines personalized HRI interfaces with intelligent robotics and IoT techniques. This proposed open-source framework demonstrates characteristics of high security, compatibility, and modularity, allowing unlimited user access. Two case studies were conducted to evaluate the proposed framework's functionalities, evaluating its effectiveness in home-care scenarios. Users' feedback was collected via questionnaires, which indicates the IoHRT framework's high potential for home-care applications.  
For project videos and tutorials, please check our website:  \url{https://sites.google.com/view/iohirt} Codes will be publicly available after this paper is accepted. 
\end{abstract}

\section{Introduction}

  Many countries worldwide are encountering significant challenges associated with their aging populations \cite{abou2020systematic}. 
 As the population decreases and the number of elderly people increases, society needs to invest a larger portion of resources to care for the elderly and individuals with disabilities, which places a heavier burden on society \cite{han2020aging}.  Home-care robots are emerging as a vital component of the assistive technology revolution, which represents a critical step toward addressing the challenges faced by aging societies \cite{gao2021progress}. Therefore, there is an urgent need to develop intelligent home-care robots to improve the quality of care for those in need and enable individuals to maintain their independence and dignity \cite{yamazaki2012home}.


In recent years,  Internet of Things (IoT) has been developed to facilitate efficient Machine to Machine (M2M) communication using standard network protocols \cite{batth2018internet}. Many countries have adopted smart home-care systems using  IoT and other technologies \cite{liu2022research}. However, the emergence of  home-care robots  drives the need for the development of advanced architectures that can integrate traditional IoT architecture with autonomous robotic systems and advanced human-robot interaction (HRI) techniques. This is due to the fact that robots are required to work collaboratively with humans in shared workspaces, which necessitates an architecture that can support ergonomic HRI with high efficiency \cite{zhang2021explainable}. The urgent need of  home-care robotic systems is instrumental in speeding up the growth of the Internet of Robotic Things (IoRT) with human-in-the-loop control \cite{vermesan2020internet}.


\begin{figure}[tb]
	\centering
  \captionsetup{font=footnotesize,labelsep=period}
\includegraphics[width = 0.9\hsize]{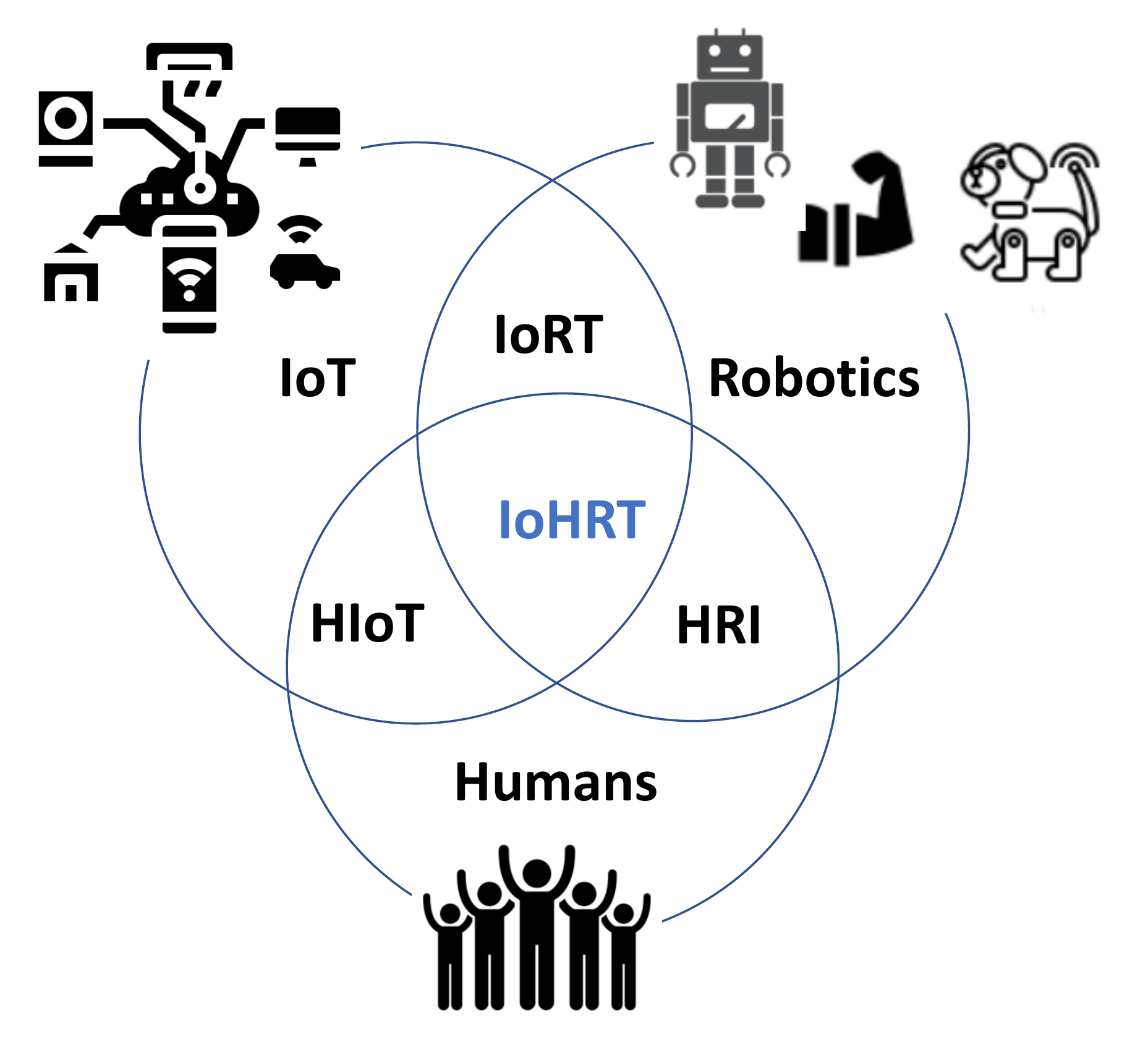}
	\caption{Illustration of the relationships of Robotics, Humans, IoT, HIoT, HRI, IoRT, and IoHRT. }
 \vspace{-0.3cm}
 \label{fig:concepts}
\end{figure}





IoRT involves robots that can monitor their surroundings, combine data from multiple sensors, utilize both local and distributed intelligence to make decisions, and take actions to manipulate physical objects. In traditional IoRT frameworks, robots normally work in structured environments to execute predefined repetitive tasks.
If the robot's operating environment changes, the robot may fail to perform its predefined tasks, resulting in time-consuming and labor-intensive redevelopment \cite{liu2020analyzing}.  Moreover, although automation has made significant progress in recent years, robots are yet to be capable of executing tasks with full autonomy in home-care applications. The integration of human control and intelligent robotics can ensure both efficiency and safety for home-care applications \cite{zhang2022human}. 
To enable caregivers to control medical or assistive robots remotely and perform home-care services with high efficiency \cite{riek2017healthcare}, HRI and teleoperation techniques are worth developing and integrating into the IoRT \cite{pang2018development, zhang2019design}. To this end, we enrich the concept of IoRT by combining the strengths of humans' cognitive skills and robots' dexterous manipulation skills for task execution, leading to the new concept of \textbf{Internet of Humans and Robotic Things (IoHRT)}.  The relationships of  Robotics, Humans, IoT, Human-centred Internet of Things (HIoT)
\cite{koreshoff2013approaching}, HRI, IoRT,  and IoHRT, are illustrated in Fig. \ref{fig:concepts}.





In addition, the recent growth of cloud infrastructures has led to the development of cloud robotics paradigms. 
 A cloud server can be used to concentrate all the resources of environmental information obtained by sensors, states of actuators, and observations from intelligent robots \cite{yang2017iot}, which enables efficient data processing and communication among different robotic platforms.  Cloud robotics is promising for  assisted living. However,  human-in-the-loop control techniques have not been integrated into most of the cloud robotic systems  for home-care applications. Therefore, we deployed the IoHRT architecture on a cloud server with a centralized hierarchy for resource management and control of multiple  robotic systems.

To our best knowledge, this paper introduces the first python-based open-source IoHRT framework for home-care applications with the support of cloud computing. The \textbf{main contributions} of this paper are listed as follows.
\begin{itemize}
    \item We proposed the concept of  IoHRT by  integrating human-in-the-loop control with traditional IoRT frameworks, which  enables robots to handle unpredictable  events effectively for home-care applications.
     \item We developed a unified framework as a ground architecture for home-care applications  and deployed the proposed framework on cloud service. We conducted experiments to prove the time delay for teleoperation is within a reasonable range.
    \item We evaluated the effectiveness of the framework with two case studies and collected users' feedback via questionnaires. We compared the proposed IoHRT with other IoT/IoRT architectures and verified its key advantages in terms of  security, compatibility, modularity, unlimited access,  and open-source nature.
\end{itemize}





The remainder of this paper is described as follows. Second II introduces the methodology for the construction of IoHRT. Case studies and system evaluation are described in Section III,  and finally, conclusions are drawn in Section IV.

\begin{figure*}[tb]
	\centering
\captionsetup{font=footnotesize,labelsep=period}
	\includegraphics[width = 0.95\hsize]{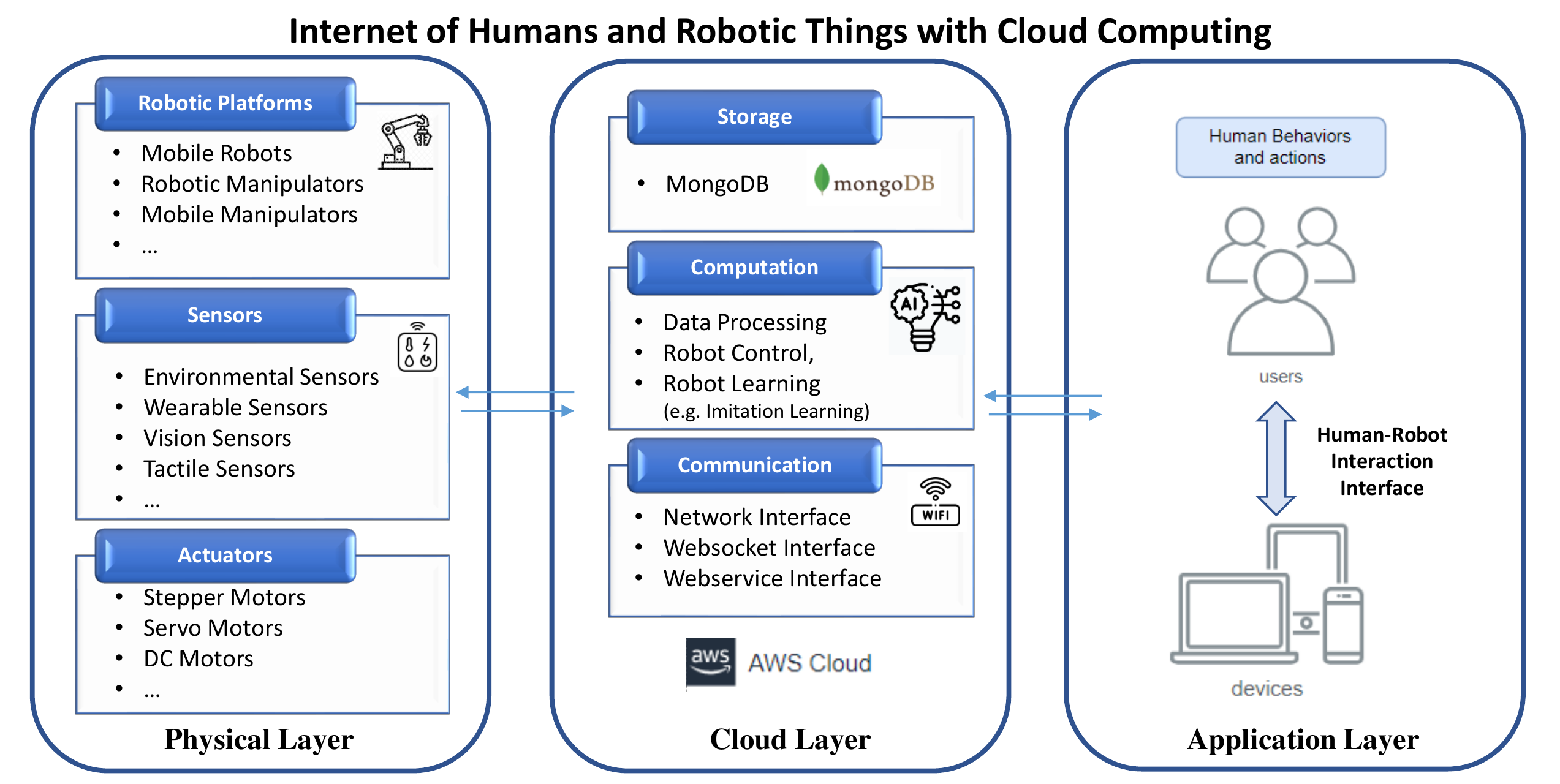}
	\caption{Overview of the IoHRT Framework with cloud computing, including key components of a physical layer, a cloud layer, and an application layer. }
  \vspace{-0.3cm}
	\label{fig:framework1}
\end{figure*}

\section{Methodology}

\subsection{Design Consideration}


We believe that several factors are hindering the use of traditional IoT/IoRT frameworks for home-care applications. 


Firstly,  most existing IoT/IoRT frameworks lack permission management functionality. Safety is the major concern for the promotion of IoHRT frameworks in home-care scenarios, where the users are mostly elderly people.  Therefore, our proposed system is designed to enable users to register with different levels of access.
The users' operation data will be stored in their own accounts to ensure security. This can also help reduce mutual influence among different users in the meantime.

Secondly, most of the existing IoT/IoRT frameworks are tailored to specific applications. This
restricts the range of tasks that robots can perform and places a heavy burden on developers who need to modify the teleoperation mode for a new task.
To address the limitations mentioned above, we aim to develop a unified framework that decouples all functions of  teleoperation systems through modular design. The signal communication protocol in the IoHRT framework enables new human interaction interfaces, mapping strategies, and robotic platforms to be added as plug-in functions for teleoperation with ease.

Thirdly, most IoRT frameworks depend on Local Area Network (LAN), failing to take advantage of the  cloud computing service. To enable remote operation and scalability, we deploy the proposed system on cloud service, which can also ensure uninterrupted and unlimited access for users.  In addition, cloud computing can accelerate AI model training and transform collected data into meaningful robot intelligence \cite{kehoe2015survey}, which can further benefit intelligent robotic platforms for home-care applications.

Finally, the design and deployment of IoRT lack open-sourced resources. To address this, we will release all the codes for IoHRT to the public. A website for this project is built, including documentation of technical details, videos, forum discussions, tutorials, etc. We encourage technicians and researchers to register for `developer mode' access, enabling easy updates of the IoHRT. The users with `developer mode' access can easily integrate open-source codes for low-level control of robotic platforms into the IoHRT, targeting at customized home-care applications. We will also record videos for professional training and share them through our Website. The proposed framework can be extended to other applications where human-in-the-loop control is necessary for interaction with robots.



Following the design considerations mentioned above, we developed the IoHRT framework in this paper. Our framework's architecture leverages broad interfaces, enabling seamless switching among different software/hardware components while preserving the software framework. This design empowers the implementation of various tasks that can run on diverse robotic platforms, which can promote the development and application of intelligent robotics in various home-care scenarios.

\subsection{Framework Construction}
Fig. \ref{fig:framework1} illustrates the philosophy and design of the IoHRT framework. The framework composes of three layers, including a physical layer (robots, sensors, actuators), a cloud layer (storage, communication, and computing), and an application layer (web service for end users).  The framework construction is detailed as follows.
\subsubsection{\textbf{Physical Layer}}

The physical layer includes assistive robots or medical robots for home-care applications, sensors for robotic perception and environmental monitoring, as well as actuators for the control of other autonomous machines.

A local edge server facilitates communication between the cloud service and multiple robotic platforms controlled by local workstations. It also manages incoming requests from the cloud.
A TCP socket communication method is used, which simplifies the integration between specific APIs for robot control and the cloud platform.

\begin{figure*}[tb]
	\centering
  \captionsetup{font=footnotesize,labelsep=period}
	\includegraphics[width = 0.95\hsize]{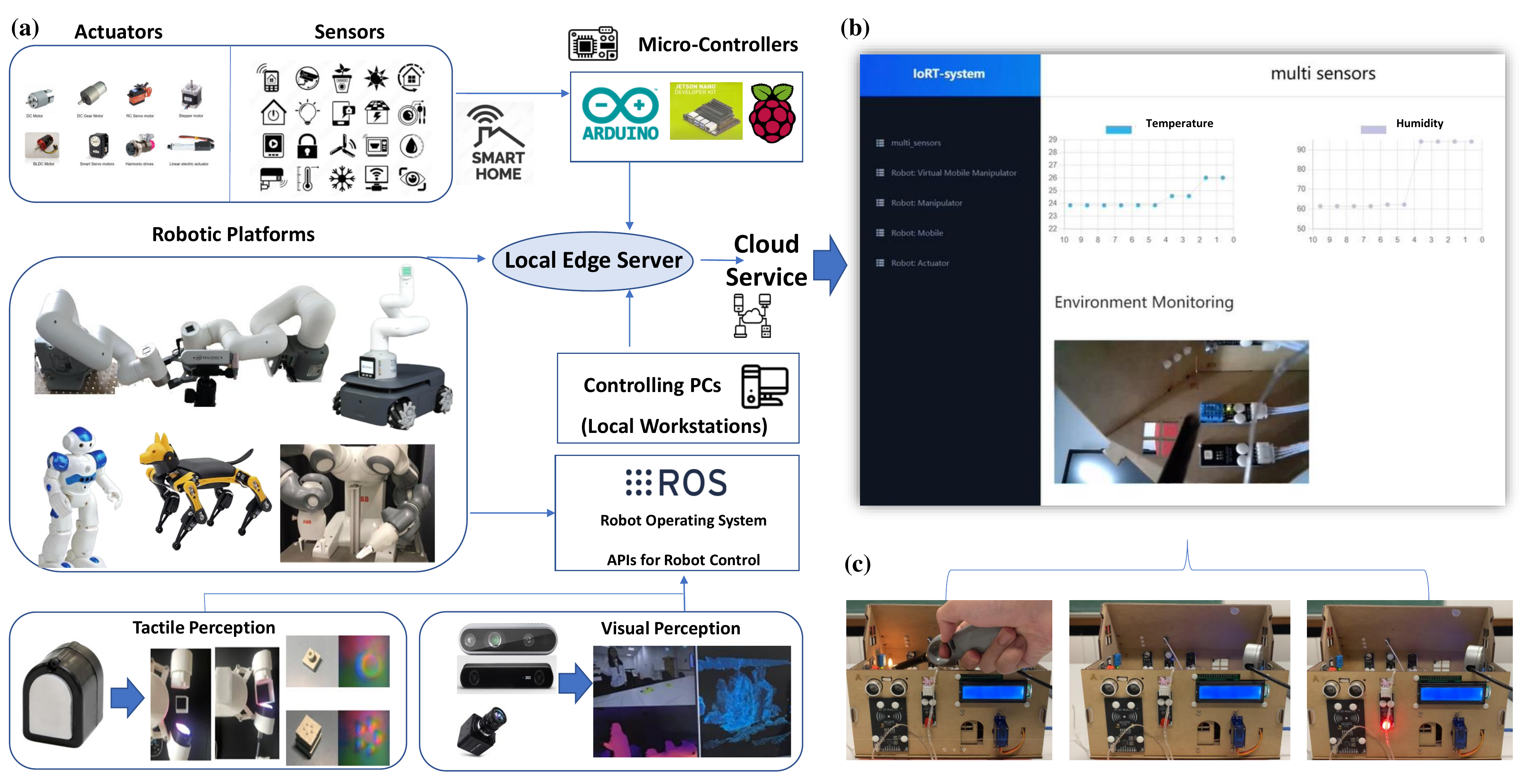}
	\caption{ Detailed exemplars of the key components in the IoHRT.  (a) Overview of the robotic things in the physical layer and the connection to cloud service. (b) Overview of the environment monitoring interface. (c) Different status of the simulated Smart Home environment. }
	\label{fig:physicA}
  \vspace{-0.3cm}
\end{figure*}

In the physical layer, different sensors' information can be captured in a real-time manner through microcontrollers, while the data can be sent to the cloud for analysis. 
 The proposed framework can enable the integration of  Smart Home \cite{babangida2022internet} and intelligent robots \cite{zhang2022one} to provide various types of home-care services. For example, temperature and humidity can be monitored, as shown in Fig. \ref{fig:physicA}(b-c). If abnormal states of the environment are discovered, the system can automatically inform security professionals to deal with emergencies. This demonstrates the fundamental functions of the IoHRT for Smart Home applications.   Fig. \ref{fig:physicA}(a) shows the other exemplars of sensors, actuators, and robots, that can be easily integrated into the physical layer of the IoHRT framework.


\subsubsection{\textbf{Cloud Layer}}
Cloud services include
web-based applications and authentication modules.
The cloud server layer can manage and arrange all the computations, storage, and communications.  The cloud server layer is developed, implemented, and deployed in one of the most prevalent commercial cloud services: AWS (Amazon Web Services). Compared to the local server, the use of cloud service has significant advantages of scalability, elasticity, agility, fault tolerance, and high availability.

\paragraph{Storage}
MongoDB, a widely used NoSQL database is adopted to store the data originating from stationary sensors, wearable sensors, actuators, and robots, as key-value pairs in the cloud. MongoDB has the properties of built-in high availability, horizontal scaling, and geographic distribution. Two types of data processing are conducted in the cloud server: real-time stream processing and post-processing. In the real-time streaming process, the cloud server processes the abnormal data and arranges an inspection mission for the robot to investigate the anomalies. Another application is teleoperation. The data sent by human operators can be transferred to the target robot for real-time control, while the data can be stored in the database for further deep learning-based studies. In post-processing, the data stored in the database are used by the computation component for analysis. 


\paragraph{Computation}
Intensive computation algorithms, including data analysis, sensor fusion, and leader-follower mapping strategies for human-in-the-loop control, are deployed in the cloud server. 
Deep learning has become a prevalent method for image classification and recognition tasks \cite{zhang2021surgical}.  It can be used to improve the perception of environmental contexts and enable robot action generation through a control policy obtained via supervised learning \cite{wang2021real}.
Most of the daily tasks, such as cooking, pouring, and cleaning, involve complex dynamic processes that are difficult to model \cite{Huang2019Accurate}.  Learning from demonstration offers a paradigm for robots to learn successful policies in fields where humans can easily demonstrate the desired behavior \cite{chen2020supervised}, while deriving mathematical models for robot control is challenging. Therefore, we adopt learning from demonstration to develop intelligent robots in the IoHRT framework, allowing for the creation of assistive robots that can adapt to different home-care scenarios and efficiently assist humans in their daily lives.


\begin{figure*}[tb]
	\centering
  \captionsetup{font=footnotesize,labelsep=period}
	\includegraphics[width = 0.95\hsize]{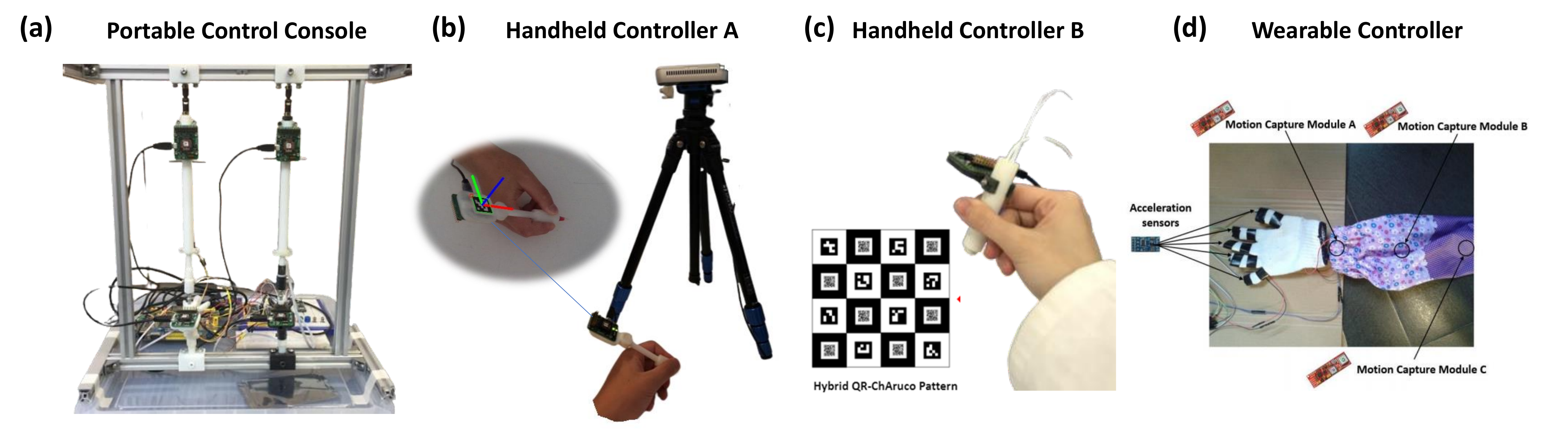}
	\caption{ Exemplars of the ergonomic portable user interfaces that can be used to control robots remotely. (a) A portable control console with IMUs for motion tracking \cite{zhang2020hamlyn}, (b) a handheld controller with Depth Camera and IMUs for motion tracking \cite{zhu2022deep}, (c) a handheld controller with Stereo Camera and IMUs for motion tracking \cite{zhang2019handheld}, (d) a Wearable controller with embedded IMUs for motion tracking.   }
	\label{fig:physicbb}
  \vspace{-0.3cm}
\end{figure*}


\paragraph{Communication}
Three communication interfaces (Network interface, Websocket interface, Web Service interface) are built to exchange messages  among robots, sensors and end-users. 
TCP socket is used to allow a two-way communication link with JSON serialized messages between robots/sensors and the servers. Websocket Interface is used for forwarding the messages received through the Network interface to the application layer, such as  temperatures, and humidities from stationary sensors. With such a socket, the end-users could  receive real-time information. The Web Service interface allows the end-users to check the data and control robots and actuators remotely with the help of the Representational state transfer (REST) API. Both Websocket and Web Service interfaces are built based on a high-level Python web framework -- Django, which enables rapid development of secure and maintainable websites. The HTML templates used for developing the  Web Service Interface are from \cite{weightking2021}.
The primary feedback system involves a webcam that observes the target object during operation. The camera captures image frames at 30Hz. 
An open-source streaming server\cite{gwuhaolin2021} with Real Time Message Protocol (RTMP)/Real Time Streaming Protocol (RTSP) has been used for video streaming. However, the high latency (about 3 seconds) prevents real-time applications, which is not suitable for robot remote control.  Considering the transmission rate of the TCP socket is slower than the UDP socket, we  use a UDP socket to transmit images obtained by cameras to the cloud service, while the images are stored in MongoDB sequentially.  Django in the cloud server retrieves the last stored frame from MongoDB and distributes it to the web client through a function called “StreamingHttpResponse()”. This lay a solid foundation for the teleoperation of robots with real-time visual feedback.

\subsubsection{\textbf{Application Layer}}
With our proposed IoHRT framework, the data obtained from stationary sensors or sensors equipped with robots can be accessed remotely by end-users.
 Based on the information obtained via data analysis, users can control the robots or actuators through a web browser by registering an account through the website. User information is stored in a built-in database in Django and authenticated by the Django authentication system during login. Each user is assigned different levels of permissions. For example, users with read-only access could only view the web page of `multi sensors'(see Fig. \ref{fig:physicA} (b)). Individuals with robot control access can operate one or more specific robots, which will be further described in the case studies section


The system can also easily integrate other types of user interaction interfaces, as long as the control commands follow a predefined protocol. For example, four types of in-house portable controllers (user interfaces) can be integrated into the IoHRT system.
Fig. \ref{fig:physicbb} (a) illustrates a foldable mechanical linkage-based user interface, which can be known as a leader robot during teleoperation. This lightweight and foldable interface employs four Inertial Measurement Units (IMUs) to estimate its end-effector pose based on forward kinematics, which can facilitate leader-follower mapping by mapping the position of the leader robot's end-effector incremental value to the follower robot's end-effector (home-care robot) after motion scaling \cite{zhang2020hamlyn}.  
Figs. \ref{fig:physicbb} (b) and (c) display different types of handheld controllers. The real-time motion tracking of the handheld controllers is implemented using vision and inertial sensor fusion \cite{zhang2019handheld, zhu2022deep}. Fig. \ref{fig:physicbb} (d) illustrates the in-house wearable controller, whose motion tracking is implemented based on accelerometers and multiple IMUs mounted at different positions of the wearable device.
More details on the user interfaces can be found in the tutorial on our website.

Each robot or actuator is arranged with a unique identification number or name so that the server can identify where the requests are sent. The robots can be repositioned remotely by human operators using different types of user interfaces, through which different types of robotic manipulation tasks can be executed with human-in-the-loop control. Additionally, live video streaming is sent to the cloud server and embedded in the web application, which allows end-users to watch live video remotely. 




\subsection{Characteristics of IoHRT}

In summary, the proposed IoHRT framework has several key characteristics that make it a useful tool for home-care applications, including:

\textbf{Security:} The framework features a security control module that is enabled by password-based authentication. All users can access the web page and view the sensors' data. Only experienced users with permission can access the robot control dashboard on the web page and manipulate robots remotely. The users with an admin role are responsible for system maintenance, while users with `developer access' can update the features of the IoHRT or integrate new robotic platforms to conduct new tasks.

 \textbf{Compatibility:} The IoHRT framework can easily integrate a variety of robotic platforms, actuators, and sensors, as long as the data structure used for information exchange follows predefined protocols. The framework is lightweight and can be executed on Windows, Mac, and Linux systems. The framework can be easily upgraded by researchers with `developer access'.

 \textbf{Modularity:} The modular design of the architecture reduces system integration efforts when changing tasks, interaction interfaces, environment perception modules, and robotic platforms. Different combinations of interaction interfaces, mapping strategies, and robotic platforms can be chosen for different applications without redesigning the teleoperation system.

\textbf{Unlimited Access:} The proposed framework can ensure uninterrupted access and location independence, allowing users to remotely control robots using any end device, such as a phone, laptop, or tablet, regardless of the platform or operating system. The system can ensure the smooth and seamless operation of different actuators/robots.

\textbf{Open-Source Nature: } The IoHRT architecture is developed based on Python, the most popular programming language for machine learning. Documentation and tutorials will be released to enable users to integrate their customized robotic platforms into the IoHRT system. The software architecture will be open-sourced to benefit the Robotics and AI community.



\section{System Evaluation}



\subsection{Case Studies}
Two case studies were conducted to evaluate the effectiveness of the proposed IoHRT framework, which demonstrates its ability to seamlessly integrate with service and medical robots. With the IoHRT system, authorized users can control specific robots with unique IDs, repositioning them to new goals or executing desired trajectories by sending control commands. They can also choose different user interfaces, modify leader-follower mapping strategies, and adjust the autonomy level of the robots for improved interaction.



\begin{figure}[tb]
	\centering
   \captionsetup{font=footnotesize,labelsep=period}
	\includegraphics[width = 1\hsize]{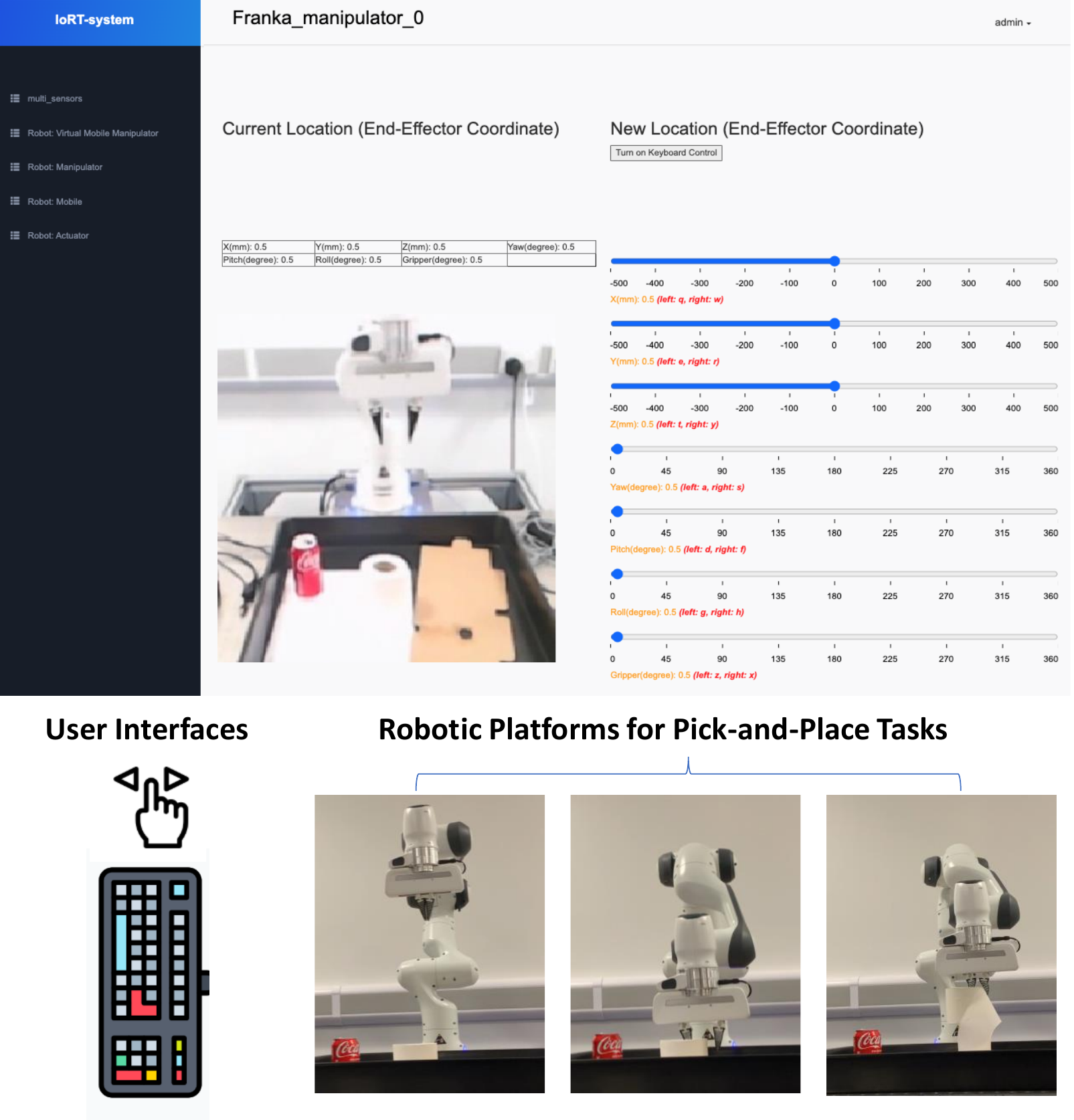}
	\caption{Overview of the experimental setup for performing the pick-and-place task in Case Study 1 using a 7-DoF robot.  }
  \vspace{-0.3cm}
	\label{fig:Franks}
\end{figure}
\subsubsection{Case Study 1 -- Pick and Place Task}


To showcase that the IoHRT framework can be used to control a robot to assist humans' daily life, we conducted the evaluation of the framework on one of the most common tasks - the pick and place task. This task involves a robotic arm approaching a target object, utilizing the gripper to pick up the object, and subsequently moving the object to a designated location.


The experimental setup is shown in Fig. \ref{fig:Franks}. A Franka
Emika Panda arm with seven joints was used for experiments. The robot arm was equipped with a soft gripper to perform the pick and place task.  A  camera was used for the robot to observe the operating scene while the real-time images were sent to the cloud server and visualized through the user interface, which provides visual feedback to human operators. The controlling PC was connected to the robot arm through Ethernet. The PC runs with a real-time kernel in Ubuntu 20 system. The robot control commands can be received from  the cloud server through TCP socket. 

Suppose that $\gamma$ is a motion scaling factor \cite{zhang2018self}, ${\Delta}t$ is the time interval for robot control. At time step $t$, $\bm{P_s(t)}$ represents the robot arm's end-effector position,  and $\bm{V_h(t)}$ represents the incremental value for control generated by humans.  $\bm{P_s(t)}$  can be determined  for robot motion control by \eqref{eq:2}
\begin{gather}
\label{eq:2}
\bm{P_s(t)}  = {\gamma}\bm{V_h(t)}{\Delta}t + \bm{P_s(t-1)}
\end{gather}

\subsubsection{Case Study 2 --  Microsurgery}

In Healthcare 4.0, a significant shift has been observed from hospital-centric healthcare to home-centric healthcare \cite{aceto2020industry}. Robotics technology can significantly benefit home-centric healthcare \cite{zhang2020automatic}. In this case, a microsurgical robot is used to demonstrate the potential of using the IoHRT Framework for  eye surgery, which can be performed by a surgeon remotely \cite{zhang2020microsurgical}.

\begin{figure}[tb]
	\centering
   \captionsetup{font=footnotesize,labelsep=period}
	\includegraphics[width = 1\hsize]{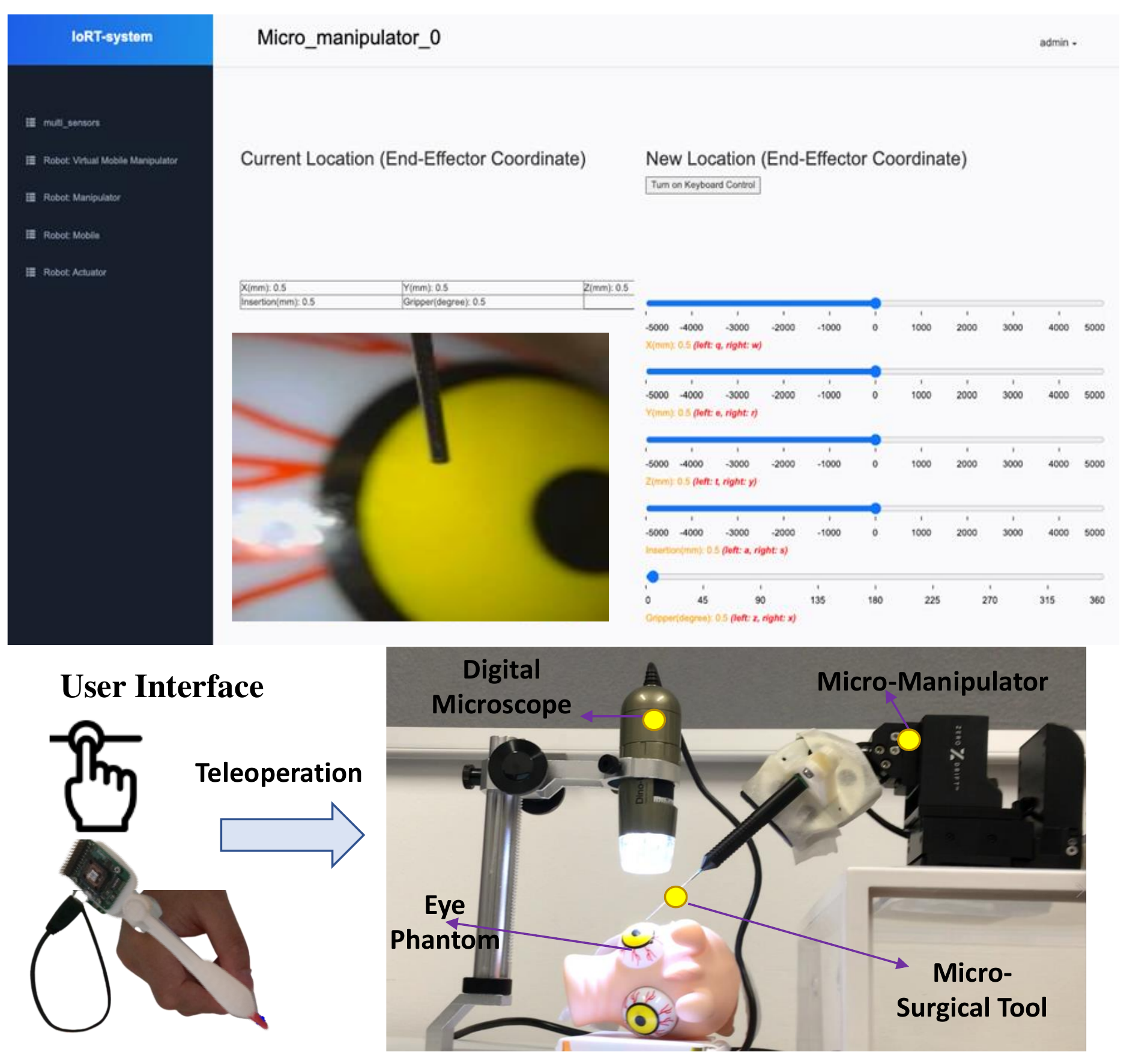}
	\caption{Overview of the scenario for the microsurgery  in Case Study 2.  An ergonomic handheld controller can be used for teleoperation. The details can be found in \cite{zhu2022deep}.}
  \vspace{-0.3cm}
	\label{fig:microrobot}
\end{figure}

Fig. \ref{fig:microrobot} illustrates the user interface and the microsurgical platform for eye surgery. The microsurgical robot is developed based on a 4-DoF positioning stage, while a digital microscope is used to provide real-time visual feedback. Different types of microsurgical tools (such as micro-needles, micro-forceps, etc) can be mounted on the microsurgical platform to perform different micro-surgical tasks.

 The control relationship between the commands generated by human operators and the microsurgical robot's movements for teleoperation is the same as  \eqref{eq:2}, while $\gamma$ can be modified based on user preference. As for human-robot shared control\cite{payne2021shared}, suppose that $\bm{V_r(t)}$ is the incremental value for control planned by the robot,  $\bm{P_s(t)}$  can be calculated by \eqref{eq:3} for motion control.
\begin{gather}
\label{eq:3}
\bm{P_s(t)}  = [(1-{m}){\gamma}\bm{V_h(t)} + {m}\bm{V_r(t)}] {\Delta}t + \bm{P_s(t-1)}
\end{gather}
where $m{\in}[0,1]$ is a weight parameter determined by the relative significance of control commands generated by humans and robots respectively.  $m = 0$ is used for teleoperation, $m = 1$ is used for fully autonomous task execution, while $m {\in}(0,1)$ is used for human-robot shared control \cite{zhang2022human}. The user can select different autonomy levels by tuning $m$.

\subsection{Performance Evaluation}

\subsubsection{Qualitative Evaluation}
We developed a questionnaire to collect user feedback on the usability of the proposed system. The user studies were conducted with ethical approval obtained from the University of Bristol Ethics Committee (ref No.10389). Eight participants, including 4 females and 4 males, were asked to complete the questionnaire.

After each experiment, participants were asked to complete a questionnaire consisting of two questions with scores ranging from 1 to 5 and an open-ended question. Higher scores indicated greater satisfaction with the system. 
The questions are listed below:
\begin{itemize}
    \item Q1:  To what extent do you think the IoHRT framework is user-friendly?
     \item Q2: How useful is the IoHRT framework for home-care applications?
     \item Q3:  Why do you feel IoRT is useful and What are the limitations of the current IoHRT framework?
\end{itemize}

\begin{figure}[tb]
	\centering
  \captionsetup{font=footnotesize,labelsep=period}
	\includegraphics[width = 0.9\hsize]{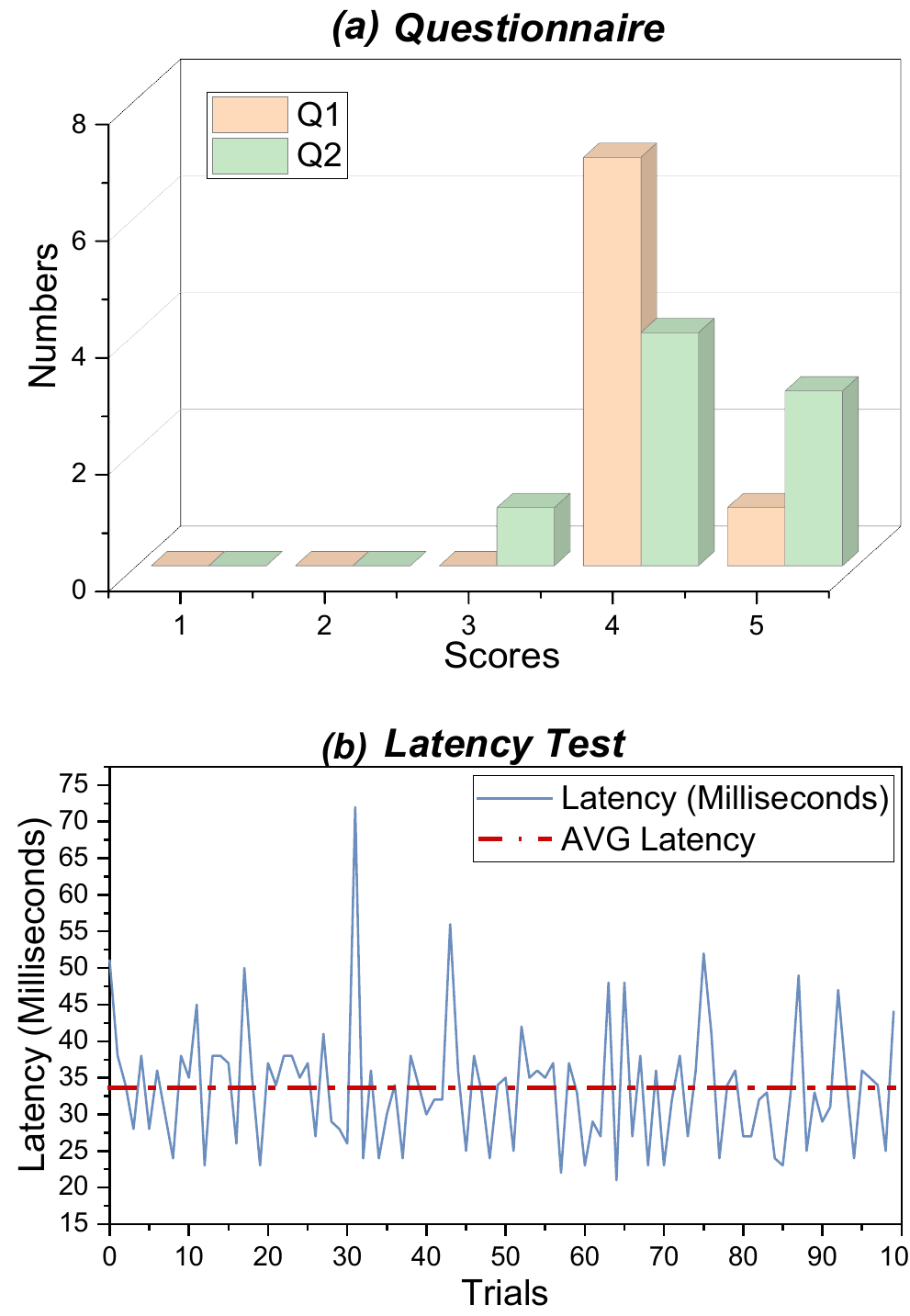}
	\caption{ Qualitative and quantitative evaluation. (a) Results of the questionnaire. (b) Results of 100 latency tests for quantitative evaluation.  }
      \vspace{-0.5cm}
	\label{fig:qs}
\end{figure}

The results of the questionnaire for two sub-tasks are shown in Fig. \ref{fig:qs} (a).  All the participants agreed that the IoHRT is user-friendly (87.5\% agreed, 12.5\% strongly agreed). With the support of the framework, they can easily interact with the robot for home-care applications.  The participants commented that the system was intuitive to use since they can easily register for an account and has access to the system with ease.  They can select robots using the GUI on the web page, and control the pose and position of the robots.   During each test, they can receive feedback with a neglectable time delay when they generate control commands to actuate the robots.    When applying the control framework to another physical robot, the user only needs to switch the robot ID to reproduce the results without  changing any codes. 87.5\%  of the participants felt that the IoHRT was helpful for home-care applications, including 37.5\% of the participants who commented that the proposed system was extremely helpful. 
The users agreed that the modular design decoupled all functions and modules of a teleoperation system, allowing researchers to select different interaction interfaces \cite{zhang2019handheld}, mapping strategies\cite{zhang2018self}, and robotic platforms based on their preferences. The inherent modularity of our system allows for seamless integration and enhances the overall flexibility of the system.


However, limitations of the current IoHRT framework were observed in the meantime. For example, the teleoperation efficiency depends on the viewpoint of the camera. At certain angles, operators may experience difficulty in performing precise manipulations due to visual-motor misalignments. More efforts should be investigated to further improve the ergonomics of human-in-the-loop control. We also received constructive feedback, such as integrating state-of-the-art Virtual Reality (VR) techniques to provide an immersive manipulation experience to the users \cite{tastan2022using}. All the discussions related to this project can be found on our website (see `Forum' subpage). We will keep improving the framework in the near future.


\subsubsection{Quantitative Evaluation}
The server is located in Virginia, US, while the client is located in Bristol, UK. The  network latency  is about 33.7ms, based on the average value obtained after 100 tests (see Fig. \ref{fig:qs} (b)). The maximum latency observed during the tests is 72ms.  The average latency for camera frame transmission is 38.8ms when using the UDP protocol. The average latency for the TCP socket is 60.0ms.

\subsection{Comparisons with Related Work}

Various IoT-enabled robotic systems have been developed for home-care applications. For example, the ABB Yumi robot has been used in teleoperation mode with wearable inertial sensors for remote control in home-centric healthcare applications \cite{zhou2019iot}. Another IoRT-based nursing home system leverages wearable and environmental sensors to detect behavioral disturbances in persons with dementia and can trigger personalized interventions by a Zora robot \cite{simoens2016internet}. In response to the COVID-19 pandemic, an IoRT system has been developed to alleviate the workload of medical and nursing staff, where robots can collect medical data and upload it to the cloud for remote monitoring \cite{9329310}. Additionally, a cloud robotic framework called C2TAM has been proposed for SLAM in heterogeneous robotic systems, enabling data sharing and easy access to information for various tasks \cite{riazuelo2014c2tam}.

\begin{table}[!htb]
	\centering
\captionsetup{font=footnotesize,labelsep=period}
	\caption{Comparisons With Related Work:``$\checkmark$`` and ``X`` indicate with and without the relevant properties, respectively. }
	\label{tab:Results-compare}
	\begin{tabular}{c|c|c|c|c|c}
		\hline\hline
	\multirow{ 1}{*}{\textbf{Features}} 	 &\multirow{ 1}{*}{\textbf{IoHRT}} & \multirow{ 1}{*}{\cite{zhou2019iot}} &\cite{simoens2016internet}&\cite{9329310}&\cite{riazuelo2014c2tam}\\\hline
Security&  $\checkmark$ & X & X &  $\checkmark$ & $\checkmark$\\
Compatibility &  $\checkmark$  & X & X & X & X \\
Modularity &  $\checkmark$  & X & X & X & X \\
Unlimited Access &  $\checkmark$ &  X  & X & $\checkmark$ & $\checkmark$\\
Open-Source Nature &  $\checkmark$ & X & X & X & X\\\hline
\textbf{HRI Techniques }& $\checkmark$ & $\checkmark$ & $\checkmark$ & X & X\\\hline
\textbf{Cloud Server }& $\checkmark$ & X & X & X &$\checkmark$ \\
\hline\hline
	\end{tabular}
 \vspace{-0.3cm}
\end{table}



Table~\ref{tab:Results-compare} compares the proposed framework with other related works that are applicable to home-care applications. The table highlights key features of existing frameworks and indicates whether they incorporate HRI technologies or cloud servers in their deployment.

\section{Discussions and Future Work}

%
 

In this paper, we propose a unified IoHRT framework that integrates HRI into traditional IoRT frameworks with cloud computing for home-care applications. The proposed IoHRT framework enables robots to share and disseminate information and allows human operators to interact with robots using ergonomic user interfaces. To evaluate the effectiveness of the proposed IoHRT framework, we demonstrate successful applications through case studies, including a pick-and-place task using a service robot and microsurgery using a microsurgical robot. User studies indicated that the proposed IoHRT framework has  evident benefits for home-care applications, due to its characteristics of Security, Compatibility, Modularity, Unlimited access, and Open-Source Nature.

In the future, we plan to integrate an adaptive cognitive module into the proposed framework, which can enable robots to detect changes in users' habits and intentions \cite{zhang2018self}, thereby providing better service to end-users. Additionally, we aim to further enhance the ergonomics \cite{zhang2020ergonomic} and the reproducibility of the proposed system by verifying the functionality of a specific robot in a simulated environment before task execution in the real environment. We also plan to integrate VR technologies to implement immersive robotic manipulation and enable operators with robot control access to practice robot operation skills through interaction with virtual avatars.



\section*{Acknowledgements}
The authors would like to acknowledge all participants who were involved in the user studies.




\bibliographystyle{IEEEtran}

\bibliography{IEEEabrv,ref}

\end{document}